# Concerning *Olga*, the Beautiful Little Street Dancer

## (Adjectives as Higher-Order Polymorphic Functions)

WALID S. SABA

American Institutes for Research
wsaba@air.org

### Abstract

In this paper we suggest a typed compositional semantics for nominal compounds of the form [*Adj Noun*] that models adjectives as higher-order polymorphic functions, and where types are assumed to represent concepts in an ontology that reflects our commonsense view of the world and the way we talk about it in ordinary language. In addition to [*Adj Noun*] compounds our proposal seems also to suggest a plausible explanation for well known adjective ordering restrictions.

## Introduction & Overview

The sentence in (1) could be uttered by someone who believes that: (*i*) Olga is a dancer and a beautiful person; or (*ii*) Olga is beautiful as a dancer (i.e., Olga is a dancer and she dances beautifully).

*Olga is a beautiful dancer* (1)

As suggested by Larson (1998), there are two possible routes to explain this ambiguity: one could assume that a noun such as 'dancer' is a simple one place predicate of type $\langle e,t \rangle$ and 'blame' this ambiguity on the adjective; alternatively, one could assume that the adjective is a simple one place predicate and blame the ambiguity on some sort of complexity in the structure of the head noun (Larson calls these alternatives *A-analysis* and *N-analysis*, respectively).

In an *A-analysis*, an approach predominantly advocated by Siegel (1976), adjectives are assumed to belong to two classes, termed predicative and attributive, where predicative adjectives (e.g. *red*, *small*, etc.) are taken to be simple functions from entities to truth-values, and are extensional, and thus intersective: 〚*Adj Noun*〛 = 〚*Adj*〛 ∩ 〚*Noun*〛. Attributive adjectives (e.g., *former*, *previous*, *rightful*, etc.), on the other hand, are functions from common noun denotations to common noun denotations – i.e., they are predicate modifiers of type $\langle\langle e,t\rangle,\langle e,t\rangle\rangle$, and are thus intensional and non-intersective (but are subsective: 〚*Adj Noun*〛 ⊆ 〚*Noun*〛). On this view, the ambiguity in (1) is explained by posting two distinct lexemes (*beautiful*$_1$ and *beautiful*$_2$) for the adjective *beautiful*, one of which is an attributive while the other is a predicative adjective. In keeping with Montague's (1970) edict that similar syntactic categories must have the same semantic type, for this proposal to work, all adjectives are initially assigned the type $\langle\langle e,t\rangle,\langle e,t\rangle\rangle$ where intersective adjectives are considered to be subtypes obtained by triggering an appropriate meaning postulate. For example, assuming the lexeme *beautiful*$_1$ is marked (e.g. by a lexical feature such as +INTERSECTIVE), the meaning postulate $\exists P \forall Q \forall x [\text{BEAUTIFUL}(Q)(x) \leftrightarrow P(x) \land Q(x)]$ would then yield an intersective meaning when $P$ is *beautiful*$_1$; and where a phrase such as 'a beautiful dancer' is interpreted as follows[1]:

〚*a beautiful*$_1$ *dancer*〛
⇒ $\lambda P[(\exists x)(\text{DANCER}(x) \land \text{BEAUTIFUL}(x) \land P(x))]$
〚*a beautiful*$_2$ *dancer*〛
⇒ $\lambda P[(\exists x)(\text{BEAUTIFUL}(\hat{}\text{DANCER}(x)) \land P(x))]$

While it does explain the ambiguity in (1), several reservations have been raised regarding this proposal. As Larson (1995; 1998) notes, this approach entails considerable duplication in the lexicon as this means that there are 'doublets' for all adjectives that can be ambiguous between an intersective and a non-intersective meaning. Another objection, raised by McNally and Boleda (2004), is that in an *A-analysis* there are no obvious ways of determining the context in which a certain adjective can be considered intersective. For example, they suggest that the most natural reading of (2) is the one where *beautiful* is describing Olga's dancing, although it does not modify any noun and is thus wrongly considered intersective by modifying Olga.

---

[1] Note that as an alternative to meaning postulates that specialize intersective adjectives to $\langle e,t\rangle$, one can perform a type-lifting operation from $\langle e,t\rangle$ to $\langle\langle e,t\rangle,\langle e,t\rangle\rangle$ (see Partee, 2007).

*Look at Olga dance. She is beautiful.* (2)

While valid in other contexts, in our opinion this observation does not necessarily hold in this specific example since the resolution of 'she' must ultimately consider all entities in the discourse, including, presumably, the dancing *activity* that would be introduced by a Davidsonian representation of 'Look at Olga dance' (this issue is discussed further below).

A more promising alternative to the *A-analysis* of the ambiguity in (1) has been proposed by Larson (1995, 1998), who suggests that 'beautiful' in (1) is a simple intersective adjective of type $\langle e, t \rangle$ and that the source of the ambiguity is due to a complexity in the structure of the head noun. More specifically, Larson suggests that a deverbal noun such as *dancer* should have a Davidsonian representation such as $(\forall x)(\text{DANCER}(x) =_{df} (\exists e)(\text{DANCING}(e) \land \text{AGENT}(e,x)))$; that is, any $x$ is a DANCER iff $x$ is the agent of some dancing activity (Larson's notation is slightly different). In this analysis, the ambiguity in (1) is attributed to an ambiguity in what 'beautiful' is modifying, in that it could be said of Olga or her dancing activity. That is, (1) is to be interpreted as follows:

$[\![Olga\ is\ a\ beautiful\ dancer]\!]$
$\Rightarrow (\exists e)(\text{DANCING}(e) \land \text{AGENT}(e, olga)$
$\quad \land (\text{BEAUTIFUL}(e) \lor \text{BEAUTIFUL}(olga)))$

In our opinion, Larson's proposal is plausible on several grounds. First, in Larson's *N-analysis* there is no need for impromptu introduction of a considerable amount of lexical ambiguity. Second, and for reasons that are beyond the ambiguity of *beautiful* in (1), there is ample evidence that the structure of a deverbal noun such as 'dancer' must admit a reference to an abstract object, namely a dancing activity; as, for example, in the resolution of 'that' in (3).

*Olga is an old dancer.* (3)
*She has been doing that for 30 years.*

Furthermore, and in addition to a plausible explanation of the ambiguity in (1), Larson's proposal seems to provide a plausible explanation for why 'old' in (4a) seems to be ambiguous while the same is not true of 'elderly' in (4b): 'old' could be said of Olga or her teaching; while 'elderly' is not an adjective that is ordinarily said of objects that are of type *activity*:

a. *Olga is an old teacher* (4)
b. *Olga is an elderly teacher*

With all its apparent appeal, however, Larson's proposal is still lacking. For one thing, and while it presupposes that some sort of type matching is what ultimately results in rejecting the subsective meaning of 'elderly' in (4b), the details of such processes are more involved than Larson's proposal seems to imply. For example, while it explains the ambiguity of 'beautiful' in (1), it is not quite clear how an *N-Analysis* can explain why 'beautiful' does not seem to admit a subsective meaning in (5).

*Olga is a beautiful young street dancer* (5)

In fact, 'beautiful' in (5) seems to be modifying Olga for the same reason the sentence in (6a) seems to be more natural than that in (6b).

a. *Maria is a clever young girl* (6)
b. *Maria is a young clever girl*

The sentences in (6) exemplify what is known in the literature as adjective ordering restrictions (AORs). However, despite numerous studies of AORs (e.g., see Wulff, 2003; Teodorescu, 2006), the slightly differing AORs that have been suggested in the literature have never been formally justified. What we hope to demonstrate below however is that the apparent ambiguity of some adjectives and adjective-ordering restrictions are both related to the nature of the ontological categories that these adjectives apply to in ordinary spoken language.

Thus, and while the general assumptions in Larson's (1995; 1998) *N-Analysis* seem to be valid, it will be demonstrated here that nominal modification seem to be more involved than has been suggested thus far. In particular, it seems that attaining a proper semantics for nominal modification requires a much richer type system than currently employed in formal semantics. Before we proceed to nominal modification, therefore, in the next section we will briefly introduce a type system that is akin to that suggested several years ago by Sommers (1963); a system that forms the foundation of a compositional semantics that is grounded in an ontology that in turn reflects our commonsense view of the world and the way we talk about it in ordinary language.

## Ontological Concepts as Types

We assume a Platonic universe that includes everything that can be spoken about in ordinary language, in a manner akin to that suggested by Hobbs (1985). However, in our formalism concepts belong to two quite distinct categories: (*i*) ontological concepts, such as `animal`, `substance`, `entity`, `artifact`, `event`, `state`, etc., which are assumed to exist in a subsumption hierarchy, and where the fact that an object of type `human` is ultimately an object of type `entity` is expressed as `human ⊑ entity`; and (*ii*) logical concepts, which are the properties (that can be said) of and the relations (that can hold) between ontological concepts. Since adjectives are our immediate concern, consider the following illustrating the difference between ontological and logical concepts:

a. DEDICATED$(x :: \texttt{human})$ (7)
b. CLEVER$(x :: \texttt{animal})$
c. IMMINENT$(x :: \texttt{event})$
d. OLD$(x :: \texttt{entity})$
f. BEAUTIFUL$(x :: \texttt{entity})$

These predicates are supposed to reflect the fact that, in ordinary spoken language, DEDICATED is a property that is ordinarily said of objects that must be of type $\texttt{human}$ (7a); that CLEVER could be said of objects of type $\texttt{animal}$ (7b); IMMINENT is a property that is said of objects that must be of type $\texttt{event}$ (7c); etc.

In addition to logical and ontological concepts, there are also proper nouns, which are the names of objects; objects that could be of any type. A proper noun, such as *sheba*, is interpreted as

$$[\![sheba]\!] \Rightarrow \lambda P[(\exists^1 x)(\text{NOO}(x :: \texttt{thing}, \text{`sheba'}) \wedge P(x :: \texttt{t}))]$$

where $\text{NOO}(x :: \texttt{thing}, s)$ is true of some unique object $x$ (which could be any $\texttt{thing}$), and $s$ if (the label) $s$ is the name of $x$, and $\texttt{t}$ is presumably the type of objects that $P$ applies to (to simplify notation we often write $[\![sheba]\!] \Rightarrow \lambda P[(\exists^1 sheba :: \texttt{thing})(P(sheba :: \texttt{t}))]$). Consider now the following, where we have assumed that THIEF$(x :: \texttt{human})$, i.e., that THIEF is a property that is ordinarily said of objects that must be of type $\texttt{human}$, and where $\mathbf{BE}(x,y)$ is true when $x$ and $y$ are the same objects[2]:

(8) $[\![sheba\ is\ a\ thief]\!]$
$\Rightarrow (\exists^1 sheba :: \texttt{thing})(\exists x)$
$(\text{THIEF}(x :: \texttt{human}) \wedge \mathbf{BE}(sheba, x))$

That is, there is a unique object named *sheba* (which could be any $\texttt{thing}$) and some $x$ such that $x$ (which must be of type $\texttt{human}$) is a THIEF and such that *sheba* is that $x$. Note now that *sheba* is associated with more than one type in a single scope, and this necessitates a type unification, where a type unification ($\texttt{s} \bullet \texttt{t}$) between two types $\texttt{s}$ and $\texttt{t}$, and where $Q \in \{\exists, \forall\}$, is defined (for now) as follows:

$$(Qx :: (\texttt{s} \bullet \texttt{t}))(P(x)) \equiv \begin{cases} (Qx :: \texttt{s})(P(x)), & if\ (\texttt{s} \sqsubseteq \texttt{t}) \\ (Qx :: \texttt{t})(P(x)), & if\ (\texttt{t} \sqsubseteq \texttt{s}) \\ (Qx :: \bot)(P(x)), & otherwise \end{cases}$$
(9)

and where
$P(x :: \bot) = \bot$
$(\texttt{t} \bullet \bot) = (\bot \bullet \texttt{t}) = \bot$

---
[2] We are using the fact that, when $a$ is a constant and $P$ is a predicate, $Pa \equiv \exists x[Px \wedge (x = a)]$ (see Gaskin, 1995).

Since $(\texttt{human} \sqsubseteq \texttt{thing})$, the type unification required in (8) now proceeds as follows:

$[\![sheba\ is\ a\ thief]\!]$
$\Rightarrow (\exists^1 sheba :: (\texttt{human} \bullet \texttt{thing}))$
$\quad (\exists x)(\text{THIEF}(x) \wedge \mathbf{BE}(sheba, x))$
$\Rightarrow (\exists^1 sheba :: \texttt{human})(\exists x)(\text{THIEF}(x) \wedge \mathbf{BE}(sheba, x))$

Finally, and since $\mathbf{BE}(sheba, x)$, we could replace $x$ by the constant *sheba* obtaining the following:

$[\![sheba\ is\ a\ thief]\!]$
$\Rightarrow (\exists^1 sheba :: \texttt{human})(\exists sheba)$
$\quad (\text{THIEF}(sheba) \wedge \mathbf{BE}(sheba, sheba))$
$\Rightarrow (\exists^1 sheba :: \texttt{human})(\exists sheba)(\text{THIEF}(sheba) \wedge True)$
$\Rightarrow (\exists^1 sheba :: \texttt{human})(\text{THIEF}(sheba))$

In the final analysis, therefore, 'Sheba is a thief' is interpreted as follows: there is a unique object named *sheba*, an object that must be of type $\texttt{human}$, and such that *sheba* is a thief[3].

Finally, note the clear distinction between ontological concepts (such as $\texttt{human}$), which Cocchiarella (2001) calls first-intension concepts, and logical (or second-intension) concepts, such as THIEF$(x)$. That is, what ontologically exist are objects of type $\texttt{human}$, not thieves, and THIEF is a mere property that we have come to use to talk of objects of type $\texttt{human}$. Moreover, logical concepts such as THIEF are assumed to be defined by virtue of some logical expression, such as $(\forall x :: \texttt{human})(\text{THIEF}(x) \equiv_{df} \varphi)$, where the exact nature of $\varphi$ might very well be susceptible to temporal, cultural, and other contextual factors, depending on what, at a certain point in time, a certain community considers an THIEF to be. What is of particular interest to us here is that logical concepts such as THIEF (or DANCER, WRITER, etc.), are defined by logical expressions that admit abstract objects such as activities, processes, states, etc., each of which could be the object of modification.

## Types and Nominal Modification

First we will use the type system described above and the notion of type unification to properly formalize the intuitions behind Larson's proposal for nominal modification. Subsequently we show that our formalism explains the relationship between the notion of intersective vs. non-intersective adjectives and that of adjective-ordering restrictions.

---
[3] The observant reader might have noticed that the removal of $\mathbf{BE}(sheba, x)$ essentially means that the copular 'is' was in this case interpreted as the 'is' of identity. This was due to the fact that in this case a subsumption relation exists between the types of the relevant objects. In other contexts, such as 'Liz is aging', 'Sheba is angry', etc., where it seems that we are dealing with the 'is' of predication, removing $\mathbf{BE}(x,y)$ involves introducing some implicit relation between the different types that do not unify ($\texttt{human}/\texttt{process}$, $\texttt{human}/\texttt{state}$), essentially resulting in interpretations such as 'Liz IS-GOING-THROUGH-THE-PROCESS-OF aging', 'Sheba IS-IN-A-STATE-OF anger', etc. Such details however are beyond the scope of this paper.

**Formalizing Larson's Proposal.** First let us begin by showing that the apparent ambiguity of an adjective such as 'beautiful' is essentially due to the fact that beautiful applies to a very generic type that subsumes many others. Consider the following, where we assume BEAUTIFUL$(x :: \texttt{entity})$; that is that BEAUTIFUL can be said of any entity:

⟦*Olga is a beautiful dancer*⟧
$\Rightarrow (\exists e :: \texttt{activity})(\exists olga :: \texttt{human})(\text{DANCING}(e) \land$
$\quad \text{AGENT}(e, olga :: \texttt{human}) \land (\text{BEAUTIFUL}(e :: \texttt{entity})$
$\quad\quad \lor \text{BEAUTIFUL}(olga :: \texttt{entity})))$

Note now that, in a single scope, $e$ is considered to be an object of type activity as well as an object of type entity, while *Olga* is considered to be a human and an entity. This, as discussed above, requires a pair of type unifications:

⟦*Olga is a beautiful dancer*⟧
$\Rightarrow (\exists e :: \texttt{activity})(\exists olga :: \texttt{human})(\text{DANCING}(e) \land$
$\quad \text{AGENT}(e, olga :: \texttt{human}) \land$
$\quad (\text{BEAUTIFUL}(e :: (\texttt{activity} \bullet \texttt{entity}))$
$\quad\quad \lor \text{BEAUTIFUL}(olga :: (\texttt{human} \bullet \texttt{entity}))))$

Note that both types unifications will all succeed in this case since $(\texttt{activity} \bullet \texttt{entity}) = \texttt{activity}$ and $(\texttt{human} \bullet \texttt{entity}) = \texttt{human}$. The final interpretation is thus the following:

⟦*Olga is a beautiful dancer*⟧
$\Rightarrow (\exists e :: \texttt{activity})(\exists olga :: \texttt{human})$
$\quad (\text{DANCING}(e) \land \text{AGENT}(e, olga) \land$
$\quad\quad (\text{BEAUTIFUL}(e) \lor \text{BEAUTIFUL}(olga)))$

In the final analysis 'Olga is a beautiful dancer' is interpreted as: Olga is the agent of some dancing activity, and either Olga is beautiful or her dancing. However, consider now the following:

⟦*Olga is an elderly teacher*⟧
$\Rightarrow (\exists e :: \texttt{activity})(\exists olga :: \texttt{human})(\text{TEACHING}(e)$
$\quad \land \text{AGENT}(e, olga :: \texttt{human})$
$\quad \land (\text{ELDERLY}(e :: (\texttt{activity} \bullet \texttt{human}))$
$\quad\quad \lor \text{ELDERLY}(olga :: (\texttt{human} \bullet \texttt{human}))))$

While the type unification involving *Olga* is straightforward, $e$ is considered here to be an object of type activity as well as an object of type human. Since $(\texttt{activity} \bullet \texttt{human}) = \perp$ this particular type unification fails, resulting in the following:

⟦*Olga is an elderly teacher*⟧
$\Rightarrow (\exists e :: \texttt{activity})(\exists olga :: \texttt{human})(\text{TEACHING}(e)$
$\quad \land \text{AGENT}(e, olga) \land (\perp \lor \text{ELDERLY}(olga)))$
$\Rightarrow (\exists e)(\text{TEACHING}(e) \land \text{AGENT}(e, olga) \land \text{ELDERLY}(olga))$

**Adjective-Ordering Restrictions.** Consider again the logical concepts given in (7). Note that BEAUTIFUL can be said of objects of type entity, and thus it can be said of a cat, a person, a city, a movie, a dance, an island, etc. Therefore, BEAUTIFUL can be thought of as a polymorphic function that applies to objects at several levels and where the semantics of this function depend on the type of the object, as illustrated in figure 1 below[4]. Thus, and although BEAUTIFUL applies to objects of type entity, in saying 'a beautiful car', for example, the meaning of BEAUTIFUL that is accessed is that defined in the type physical (which could in principal be inherited from a supertype). Moreover, and as is well known in the theory of programming languages, one can always perform type casting upwards, but not downwards (e.g., one can always view a car as just an entity, but the converse is not true)[5].

For example, assuming RED$(x :: \texttt{physical})$ and BEAUTIFUL$(x :: \texttt{entity})$; that is, assuming that RED can be said of physical objects and BEAUTIFUL can be said of any entity, then the type casting required in (11a) is valid, while that in (11b) is not.

a. BEAUTIFUL(RED$(x :: \texttt{physical}) :: \texttt{entity}$)  (11)
b. RED(BEAUTIFUL$(x :: \texttt{entity}) :: \texttt{physical}$)

This, in fact, is precisely why 'Jon owns a beautiful red car' is more natural than 'Jon owns a red beautiful car'. In general, a sequence $\text{A}_1(\text{A}_2(x :: \texttt{s}) :: \texttt{t})$ is a valid sequence iff $(\texttt{s} \sqsubseteq \texttt{t})$. Note that this is different from type unification, in that the unification does succeed in both cases in (11). However, before we perform type unification the direction of the type casting must be valid. The importance of this interaction will become apparent below.

**How an Ambiguous Adjective Gets One Meaning.** Let us explain the example in (5), where we argued that Larson's proposal cannot explain why 'beautiful', which is considered to be ambiguous in (1), does not admit a subsective meaning in (5).

⟦*Olga is a beautiful young dancer*⟧
$\Rightarrow (\exists e :: \texttt{activity})(\exists olga :: \texttt{human})$
$\quad (\text{DANCING}(e)$
$\quad \land \text{AGENT}(e, olga :: \texttt{human})$
$\quad \land (\text{BEAUTIFUL}(\text{YOUNG}(e :: \texttt{physical}) :: \texttt{entity})$
$\quad\quad \lor \text{BEAUTIFUL}(\text{YOUNG}(olga :: \texttt{physical})$
$\quad\quad\quad :: \texttt{entity})))$

Note now that the casting required is valid in both cases. In other words, the order of adjectives is valid.

---

[4] It is perhaps worth investigating the relationship between the number of meanings of a certain adjective (say in a resource such as WordNet), and the number of **different** functions that one would expect to define in the subsumption hierarchy.
[5] Technically, the reason we can always cast up is that we can always ignore additional information. Casting down, which entails adding information, is however undecidable.

This means that we can now perform the required type unifications, which would proceed as follows:

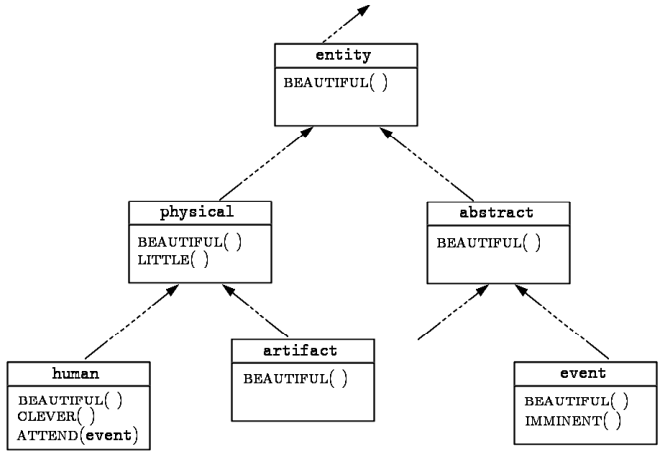

**Figure 1.** Adjectives as polymorphic functions

⟦*Olga is a beautiful young dancer*⟧
⇒ (∃e :: **activity**)(∃olga :: **human**)(DANCING(e) ∧
     AGENT(e, olga :: **human**) ∧
     (BEAUTIFUL(YOUNG(e :: (**activity** • **physical**)))
     ∨ BEAUTIFUL(YOUNG(olga
         :: (**human** • **physical**))))))

By (9), and since (**activity** • **physical**) = ⊥, the term YOUNG(e :: (**activity** • **physical**)) is reduced to YOUNG(e ::⊥) and subsequently to ⊥. Finally, and taking (⊥ ∨ β) = β we eventually get the following:

⟦*Olga is a beautiful young dancer*⟧
⇒ (∃e :: **activity**)(∃olga :: **human**)
     (DANCING(e)
     ∧ AGENT(e, olga)
     ∧ BEAUTIFUL(YOUNG(olga)))

Note here that since BEAUTIFUL was preceded by YOUNG, it could have not been applicable to an abstract object of type **activity**, but was instead reduced to that defined at the level of **physical**, and subsequently to that defined at the type **human**. A valid question that comes to mind here is how then do we express the thought 'Olga is a young dancer and she dances beautifully'. The answer is that we usually make a statement such as this:

*Olga is a young and beautiful dancer*        (12)

Note that in this case we are essentially overriding the sequential processing of the adjectives, and thus the adjective-ordering restrictions (or, equivalently, the type-casting rules!) are no more applicable. That is, (12) is essentially equivalent to two sentences that are processed in parallel:

⟦*Olga is a young and beautiful dancer*⟧
≡ ⟦*Olga is a young dancer*⟧
     ∧ ⟦*Olga is a beautiful dancer*⟧

Note now that 'beautiful' would again have an intersective and a subsective meaning, although 'young' will only apply to *Olga* due to type constraints:

⟦*Olga is a young and beautiful dancer*⟧
⇒ (∃e :: **activity**)(∃olga :: **human**)
     (DANCING(e)
     ∧ AGENT(e, olga :: **human**)
     ∧ (BEAUTIFUL(e) ∨ BEAUTIFUL(olga))
     ∧ YOUNG(olga))

## The Rich Type Structure of Nominals

That an adjective such as *beautiful* can potentially modify Olga's dancing in a sentence such as 'Olga is a beautiful dancer' is clearly due to the fact that the deverbal noun 'dancer' a complex structure that contains, at a minimum, references to a dancing **activity**, as well as the agent of the activity. A noun such as 'dancer' can however be potentially modified by other nouns, in which case the target of these modifications are some attributes of the dancing activity. For example, in

a. *Olga is a street dancer*        (13)
b. *Olga is a night dancer*
c. *Olga is a flamenco dancer*

we are clearly describing Olga's dancing **activity** by stating where she (usually) dances (13a); when she dances (13b); and what she dances (13c). It would seem therefore that a deverbal noun such as 'dancer' must have a structure such as the following:

(∀x)(DANCER(x) =_{df}
     (∃e :: **activity**)(∃x :: **human**)(∃y :: **time**)
       (∃z :: **location**)(∃u :: **content**)
         (DANCING(e)
           ∧ AGENT(e, x)
           ∧ TIME(e, y)
           ∧ LOCATION(e, z)
           ∧ THEME(e, u)
     ))

Now assuming that *street* :: **location**, *night* :: **time**, and *flamenco* :: **content** then type unification, along the lines described above, would ensure that each noun modifies the correct slot. Furthermore, there seems to be the equivalence of adjective-ordering restrictions, in that sequences of nouns usually obey some strict rules. For example, (14a) is clearly more natural than (14b), although the type casting rules seem to be more complex than in the case of adjectives.

a. *Olga is a flamenco street dancer* (14)
b. #*Olga is a street flamenco dancer*

Furthermore, it seems that these rules are also a function of the general category of the deverbal noun. For example, while 'dancer' is a noun that is derived from an activity verb, there are various subclasses within this general category that have a slightly different structure, not to mention deverbal nouns that are derived from other verb classes (e.g., process, state, etc). For example, the deverbal noun 'offer', is considerably different from the deverbal noun 'dancer' in this important respect:

$\text{DANCER}(x) \equiv x$ *is the* **agent** *of a dancing activity*
$\text{OFFER}(x) \equiv x$ *is the* **object** *of an offering activity*

Thus, and while both 'generous' and 'attractive' are adjectives that can be said of an object of type **human**, it seems that only 'generous' but not 'attractive' can be predicated of Jon in (15).

a. *Jon made a generous offer* (15)
b. *Jon made an attractive offer*

Formulating these processes in a strongly typed compositional semantics would have to however wait for another place and another time.

## Concluding Remarks

In this paper we have shown that nominal modification can be adequately treated in a semantics embedded in a strongly-typed ontology; an ontology that reflects our commonsense view of the world and the way we talk about it in ordinary language. While our concern in this paper was the semantics of [*Adj Noun*] nominals, our proposal seems to also provide an explanation for some well-known adjective-ordering restrictions and might also provide a plausible framework for the semantics of [*Noun Noun*] compounds.

## References

Cocchiarella, N. B. (2001), Logic and Ontology, *Axiomathes*, **12**, pp. 117-150.
Gaskin, R. (1995), Bradley's Regress, the Copula, and the Unity of the Proposition, *The Philosophical Quarterly*, **45** (179), pp. 161-180.
Hobbs, J. (1985), Ontological Promiscuity, In *Proc. of the 23rd Annual Meeting of the Association for Computational Linguistics*, pp. 61-69, Chicago, Illinois, 1985.
Larson, R. (1995), Olga is a Beautiful Dancer, Presented at the Winter Meetings of the Linguistic Society of American, New Orleans.
Larson, R. (1998), Events and Modification in Nominals, In D. Strolovitch and A. Lawson (Eds.), *Proceedings from Semantics and Linguistic Theory (SALT) VIII*, pp. 145-168, Ithaca, NY: Cornell University Press.
McNally, L. and Boleda, G. (2004), Relational Adjectives as Properties of Kinds, In O. Bonami and P. Cabredo Hpfherr (Eds.), *Empirical Issues in Formal Syntax and Semantics*, **5**, pp. 179-196.
Montague, R. (1970), English as a Formal Language, In R. Thomasson (Ed.), *Formal Philosophy − Selected Papers of Richard Montague*, New Haven, Yale University Press.
Partee, B. (2007), Compositionality and Coercion in Semantics − the Dynamics of Adjective Meanings, In G. Bouma et. al. (Eds.), *Cognitive Foundations of Interpretation*, Amsterdam: Royal Netherlands Academy of Arts and Sciences, pp. 145-161.
Seigel, E. (1976), *Capturing the Adjective*, Ph.D. dissertation, University of Massachusetts, Amherst.
Sommers, F. (163), Types and Ontology, *Philosophical Review*, LXXII, pp. 327-363
Teodorescu, A. (2006), Adjective ordering restrictions revisited, In D. Baumer et. al. (Eds.), *Proceedings of the 25th West Coast Conference on Formal Linguistics*, pp. 399-407, Somerville, MA.
Wulff, S. (2003), A Multifactorial Analysis of Adjective Order in English, *International Journal of Corpus Linguistics*, pp. 245-282.